\documentclass[lettersize,journal]{IEEEtran}
\usepackage{amsmath,amsfonts,amssymb}
\usepackage{algorithmic}
\usepackage{algorithm}
\usepackage{array}
\usepackage{hyperref}
\usepackage{textcomp}
\usepackage{stfloats}
\usepackage{url}
\usepackage{verbatim}
\usepackage{graphicx}
\usepackage{subfigure}
\usepackage{cite}
\usepackage{xurl}

\usepackage{tikz}  
\newcommand*\circled[1]{\tikz[baseline=(char.base)]{
            \node[shape=circle,draw,inner sep=1pt] (char) {#1};}}

\begin{document}

\title{Privacy-Preserving Personalized Federated Learning for Distributed Photovoltaic Disaggregation under Statistical Heterogeneity}

\author{Xiaolu~Chen,
        Chenghao~Huang,
        Yanru~Zhang,~\IEEEmembership{Member,~IEEE},
        and~Hao~Wang,~\IEEEmembership{Member,~IEEE}
\thanks{This work was supported in part by the Australian Research Council (ARC) Discovery Early Career Researcher Award (DECRA) under Grant DE230100046 and the Key Project of Sichuan Science and Technology Program under Grant No. 2024YFG0006 and 2024ZYD0274. (Corresponding authors: Hao Wang, Yanru Zhang.)}
\thanks{X. Chen is with the School of Computer Science and Engineering, University of Electronic Science and Technology of China, Chengdu, China (e-mail:
202222080738@std.uestc.edu.cn).}
\thanks{C. Huang and H. Wang are with the Department of Data Science and AI, Faculty of IT and Monash Energy Institute, Monash University, Melbourne, VIC 3800, Australia (e-mails: \{chenghao.huang, hao.wang2\}@monash.edu).}
\thanks{Y. Zhang is with the School of Computer Science and Engineering, University of Electronic Science and Technology of China, Chengdu, and Shenzhen Institute for Advanced Study of UESTC, Shenzhen, China (e-mail: yanruzhang@uestc.edu.cn).}
}

\maketitle

\begin{abstract}
The rapid expansion of distributed photovoltaic (PV) installations worldwide, many being behind-the-meter systems, has significantly challenged energy management and grid operations, as unobservable PV generation further complicates the supply-demand balance. Therefore, estimating this generation from net load, known as PV disaggregation, is critical.
Given privacy concerns and the need for large training datasets, federated learning becomes a promising approach, but statistical heterogeneity, arising from geographical and behavioral variations among prosumers, poses new challenges to PV disaggregation.
To overcome these challenges, a privacy-preserving distributed PV disaggregation framework is proposed using Personalized Federated Learning (PFL). 
The proposed method employs a two-level framework that combines local and global modeling. 
At the local level, a transformer-based PV disaggregation model is designed to generate solar irradiance embeddings for representing local PV conditions. A novel adaptive local aggregation mechanism is adopted to mitigate the impact of statistical heterogeneity on the local model, extracting a portion of global information that benefits the local model.
At the global level, a central server aggregates information uploaded from multiple data centers, preserving privacy while enabling cross-center knowledge sharing. 
Experiments on real-world data demonstrate the effectiveness of this proposed framework, showing improved accuracy and robustness compared to benchmark methods.
\end{abstract}


\section{Introduction}

\subsection{Background and Motivation}
The global expansion of photovoltaic (PV) installations has accelerated in recent years, especially in small-scale distributed generation systems connected to distribution networks~\cite{world2021tracking}. 
In Australia, the total capacity of small-scale solar systems has reached 24.75 GW in 2024, with 3.96 million installations~\cite{website1}. Projections estimated that the total installed capacity of PV systems will increase six-fold over 2018 levels by 2030 and surpass 8,000 GW by 2050~\cite{ERDENER2022112224}. 
Most distributed PV systems are installed Behind-The-Meter (BTM), meaning they cannot be directly monitored by utility companies.
However, the widespread deployment of BTM PV systems poses significant challenges for energy management and grid operations, as these installations introduce additional uncertainties to load forecasting and reserve power flows~\cite{9766031,wei2023model}. 
To tackle the above challenge, estimating unobservable PV generation from net load has emerged as a promising approach, called PV disaggregation.
Accurate PV disaggregation can provide useful information for energy management and grid operations.

Deep Learning (DL) has been applied to PV disaggregation, achieving reasonably high accuracy~\cite{ERDENER2022112224}. Related works will be introduced in Section~\ref{subsec:lit}.
However, centralized data-driven PV disaggregation methods raise privacy concerns, as fine-grained electricity usage data can expose private lifestyle and habits of prosumers~\cite{8291011}.
Therefore, privacy-preserving PV disaggregation becomes essential when prosumers' data cannot be centrally stored and processed.
Notably, data-driven methods often require large training datasets, whereas distributed computation frameworks eliminate the need for centralized data storage, making them an indispensable alternative.
In summary, developing a privacy-preserving, accurate, and distributed PV disaggregation framework is beneficial for utility companies to effectively monitor distributed PV generation, enhancing energy system efficiency, reliability, and safety.

Federated Learning (FL) is well suited for distributed PV disaggregation tasks. But traditional FL frameworks, e.g., FedAvg~\cite{HOSSEINI2023116900}, do not adequately account for the statistical heterogeneity inherent in PV disaggregation, which can significantly hinder model convergence and degrade overall performance \cite{zhao2018federated}. This heterogeneity generally arises from several key factors.
\begin{itemize}
    \item \textbf{Geographical Heterogeneity}: Due to regional variations in solar irradiation, the distributed PV generation varies across different regions. 
    \item \textbf{Heterogeneity of Prosumer Behavior}: Meter data is collected from regions with diverse socioeconomic conditions, living environments, and energy consumption patterns. Therefore, prosumers exhibit significant variability in electricity usage habits and PV power usage.
    \item \textbf{Data Scarcity}: When utility companies expand their operations into new regions with new customers, these areas often lack sufficient historical data. The aforementioned heterogeneity between new regions and existing ones can limit the effectiveness of PV disaggregation in new regions, in particular during the initial period.
\end{itemize}

Therefore, a privacy-preserving distributed framework is needed to address the challenges posed by the aforementioned statistical heterogeneity inherent in PV disaggregation.

\subsection{Literature Review}\label{subsec:lit}
Data-driven methods have become popular due to their ability to function without physical models, offering greater applicability in real-world problems. 
Among data-driven approaches, Machine Learning (ML) and DL methods are widely applied.
For example, Pan et al.~\cite{PAN2022118450} proposed an unsupervised learning approach to PV disaggregation considering PV conversion efficiency due to ambient temperature variation.
Model-free approaches~\cite{9247468,9477124} utilized dictionary learning techniques to learn patterns from historical datasets with partial labels.
Chen et al.~\cite{chen2025sea} developed a PV disaggregation method using multi-scale temporal feature extraction.
Saffari et al.~\cite{10155762} proposed a spatiotemporal graph sparse coding capsule network for accurate BTM load and PV generation estimation.
Dolatabadi et al.~\cite{9247187} presented a scalable, privacy-preserving distributed parallel optimization framework for managing large-scale PV-battery aggregations, employing a linear programming-based optimization approach with distributed ledger technology for privacy.
Despite the extensively-explored research of data-driven methods for PV disaggregation, these methods often require a large amount of electricity data from producers' smart meters for centrally training, raising concerns about potential privacy breaches. 

To address this issue, recent studies~\cite{HOSSEINI2023116900,9548947} have employed FL frameworks for distributed PV disaggregation.
FL is a distributed machine learning paradigm that enables multiple devices or datasets to collaboratively train a global model without sharing their local data. Thus, FL can significantly enhance privacy and reduce data transmission by keeping the data localized and only transmitting model updates, making it particularly suitable for privacy-sensitive applications.
Lin et al.~\cite{9548947} proposed a Bayesian neural network-based FL framework for probabilistic disaggregation of behind-the-meter PV generation, utilizing a layer-wise parameter aggregation strategy for FL.
Hosseini et al.~\cite{HOSSEINI2023116900} adopted FedAvg as the FL framework, where the local model is a multi-layer perceptron (MLP) without explicitly modeling temporal dependencies in PV diagregation. Moreover, FedAvg does not effectively address statistical heterogeneity, which is a key challenge in distributed PV disaggregation.
Beyond PV disaggregation, many studies have applied FL for privacy-preserving, distributed applications across various industries.
Zhang et al.~\cite{10121172} proposed FedBIP for wind turbine blade icing prediction, and Sun et al.~\cite{10371403} proposed FedAlign for machine fault diagnosis.
Wang et al.~\cite{10535211} focused on distributed PV ultra-short-term power forecasting using FL. 
These studies demonstrate the effectiveness of FL in supporting privacy-preserving and distributed model training across a range of applications.

Traditional FL faces limitations in handing heterogeneous scenarios. Personalized Federated Learning (PFL) has emerged as an effective technique for addressing statistical heterogeneity~\cite{fallah2020personalized,SABAH2024122874,9743558}, which is exactly the primary challenge in distributed PV disaggregation tasks.
Unlike traditional FL relying on a single, globally shared model, PFL enables the development of personalized local models through customized local training~\cite{9743558}. For example, Wang et al.~\cite{10711907} proposed DSHFT, a domain separation-based heterogeneous federated transfer learning approach for remaining useful life prediction of storage hard drives. Han et al.~\cite{10720106} introduced CIGPFL, a class information-guided PFL framework for gearbox fault diagnosis. Yang et al.~\cite{10666741} proposed a clustering-based PFL approach for wafer defect classification. These studies have demonstrated the capability of PFL to effectively address statistical heterogeneity in practical applications, suggesting its potential for addressing privacy-preserving distributed PV disaggregation problems.

\subsection{Main Work and Contributions}
In this paper, a privacy-preserving distributed PV disaggregation framework is proposed for PV prosumers under statistical heterogeneity.
The framework adopts the PFL paradigm, organized into local and global levels.
At the local level, there are multiple data centers located at different regions, and each data center can access meter data within its jurisdiction to train local models. The local DL model is designed using transformer-based architecture for each data center to capture complex temporal patterns and internal relationships between multiple variables, including net load and solar irradiance. 
Considering that PV generation is primarily influenced by weather conditions, particularly solar irradiance, incorporating solar irradiance data with net load can enhance the accuracy of PV disaggregation. 
Furthermore, a novel local aggregation mechanism is adopted to selectively acquire global knowledge, because statistical heterogeneity across regions can cause local knowledge bias and degrade the representational capability of the global information, thus harming PV disaggregation accuracy of each data center.
To address it, a weighting factor calculated by solar irradiance embeddings from the designed local model, dynamically adjusts the aggregation proportion of the global model parameters, since the solar irradiance embeddings encode temporal features in irradiance patterns to represent local PV conditions more effectively. 

Additionally, a model-splitting mechanism is adopted for sharing generalized knowledge while keeping personalized knowledge for each data center. Specifically, the local DL model is divided into lower and higher layers. As the lower layers are validated to capture more generalized information compared to the higher layers~\cite{NIPS2014_375c7134}, the lower layers are transmitted with the cloud server at the global level for sharing, and the higher layers are remained locally.

At the global level, the server aggregates the uploaded information from each data center to form the global model. It provides additional knowledge to each local model to enhance disaggregation performance and is sent back for local training and aggregation.

The contributions of this work are as follows.
\begin{itemize}
    \item This work addresses the PV disaggregation problem under statistical heterogeneity in a privacy-preserving distributed learning scenario. Statistical heterogeneity, arising from geographical variations in PV generation, diverse prosumer behavior, and data scarcity, presents a major challenge that needs to be addressed.
    \item A privacy-preserving distributed PV disaggregation framework is proposed based on the PFL paradigm. Specifically, a DL model based on Transformer is designed for local PV disaggregation, capturing temporal dependencies in net load features and solar irradiance features to enhance disaggregation performance. Furthermore, a novel adaptive local aggregation mechanism is adopted in the PFL framework to mitigate inter-regional statistical heterogeneity, allowing local models to selectively extract useful global information.
    \item Extensive experiments on real-world datasets demonstrate the effectiveness of the proposed approach. The results indicate that the Transformer-based local model along with the PFL training process enables high-accuracy PV disaggregation under statistical heterogeneity.
\end{itemize}

The remainder of this paper is organized as follows. Section~\ref{Sec:PS} presents the problem statement of distributed PV disaggregation in the PFL paradigm. Section~\ref{Sec:M} describes the proposed methodology, including feature engineering, the PV disaggregation model, and the adaptive PFL framework. Section~\ref{Sec:E} provides experimental results and analysis to validate the proposed method. Section~\ref{Sec:C} presents conclusions.

\section{Problem Statement}\label{Sec:PS}
\begin{figure}[t]
\centering
\includegraphics[width=0.48\textwidth]{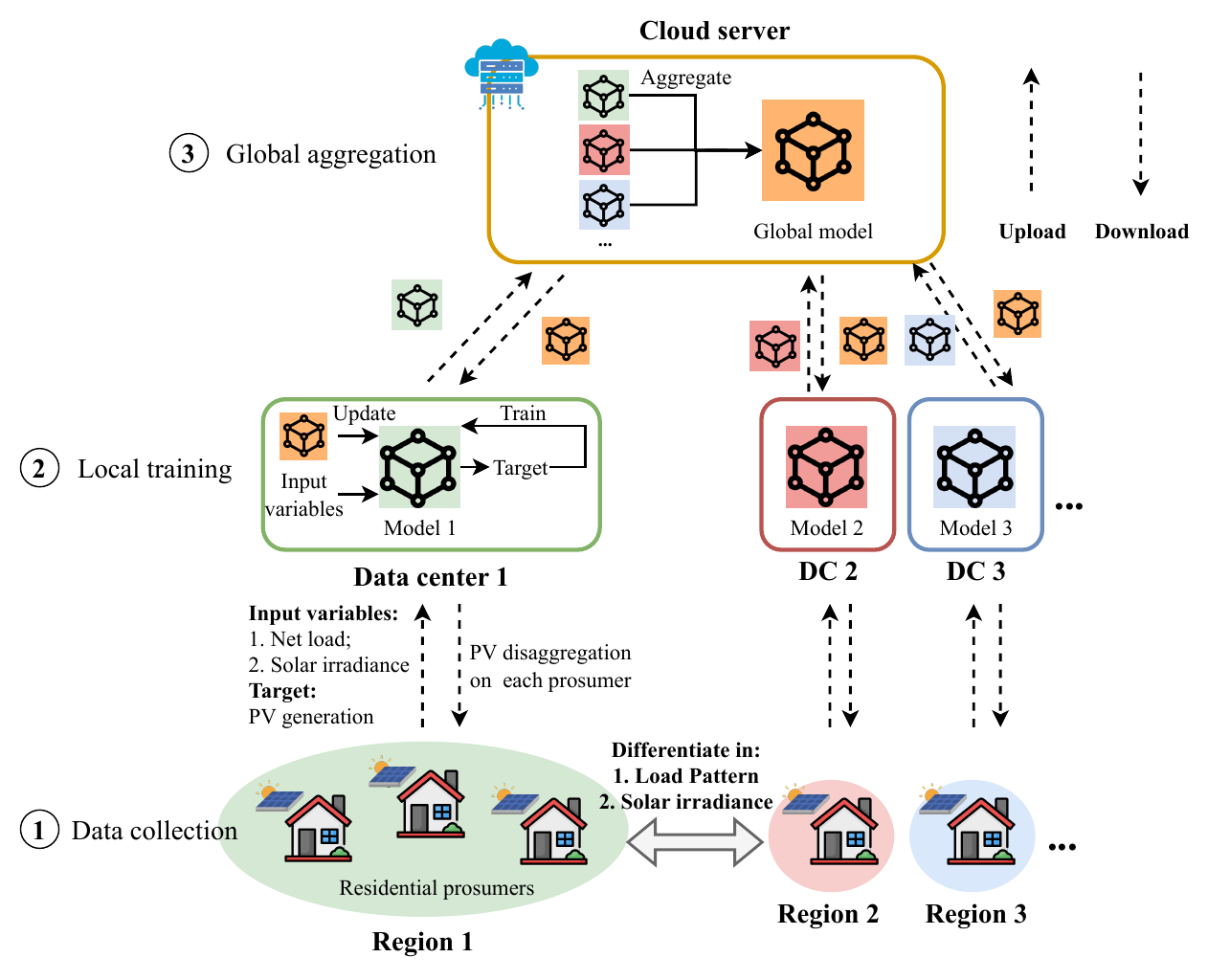}
\caption{The PFL paradigm for distributed PV disaggregation, which consists of a cloud server and multiple data centers, each serving a distinct region.}
\label{fig:system model}
\end{figure}

In this section, the fundamental concept of PV disaggregation is introduced, followed by the formulation of the problem within the PFL framework to be studied in this paper, as shown in Fig.~\ref{fig:system model}. The framework consists of a cloud server and multiple data centers, each serving a distinct region. 
There are three components of distributed PV disaggregation.
1) Data collection: Each data center collects region-specific prosumer data, including net load, solar irradiance, and PV generation, forming a private prosumer dataset. These data patterns vary across regions due to differences in geography and prosumer behavior, particularly in solar irradiance and net load.
2) Local training: Each data center performs local model training using its collected dataset. In this local training process, net load and weather data serve as input variables, while disaggregated PV generation is used as the model's output. 
After training, each data center uploads key information derived from its local model, such as model parameters or data embeddings, to the cloud server. 
3) Global aggregation: The cloud server aggregates global information using the local information received from all data centers. After completing global aggregation, the refined global information is sent back to each data center.
Subsequently, each center uses this information to enhance its local training while maintaining personalization tailored to its regional data. This iterative process of local training followed by global aggregation continues until the local models converge. Finally, the local model can be used for PV disaggregation of each prosumer in the specific region.

\subsection{PV Disaggregation}
For each day $d$ in total $D$ days, there are $T$ time slots. The net electricity load of a distributed solar prosumer at time $t$ is denoted as $x^{\text{Net},d}_t$, the corresponding PV generation as $y^{\text{PV},d}_t$, and the actual electricity consumption as $y^{\text{Actual},d}_t$ which may consist of energy from both the grid and solar generation. The relationship between these variables can be expressed as:
\begin{align}
    x^{\text{Net},d}_t = y^{\text{Actual},d}_t - y^{\text{PV},d}_t.
\end{align}

For utility companies, a portion of their prosumers have PV systems that are not BTM, meaning their smart meters record both net load $\mathbf{x}^{\text{Net},d} = \{x^{\text{Net},d}_t\}_{t=1}^T$ and PV generation $\mathbf{y}^{\text{PV},d} = \{y^{\text{PV},d}_t\}_{t=1}^T$. Consequently, utility companies have access to both consumption and generation data for this subset of prosumers. By fundamentally considering net load as training input and PV generation as the truth, the PV disaggregation task can be modeled as a supervised learning problem, where transferring training data of prosumers with PV generation readings to those without PV generation readings is important in practical applications.

Furthermore, since the PV panel characteristics of each prosumer may remain unknown, weather conditions, denoted as $\mathbf{x}^{\text{Weather},d} = \{x^{\text{Weather},d}_t\}_{t=1}^T$, should be included as assistant information for performance enhancement. Thus, for the $d$-th day, the feature space $\mathcal{X}$ consists both net load and weather conditions, represented as:
\begin{align}
    \mathcal{X} = \{[\mathbf{x}^{\text{Net},d},\mathbf{x}^{\text{Weather},d}]\}_{d=1}^D \in \mathbb{R}^{D\times 2\times T}.
\end{align}
Briefly, $[\mathbf{x}^{\text{Net},d},\mathbf{x}^{\text{Weather},d}]$ is denoted as $X^d$.
The target space $\mathcal{Y}$ contains the ground truth corresponding to the PV generation data for the same day:
\begin{align}
    \mathcal{Y} = \{\mathbf{y}^{\text{PV},d}\}_{d=1}^D \in \mathbb{R}^{D\times T}.
\end{align}
The objective of the PV generation disaggregation task is to learn a function $f(\cdot)$ with model parameters $\theta$ to achieve $f(\theta): \mathcal{X} \to \mathcal{Y}$.

\subsection{Distributed PV Disaggregation in PFL Paradigm}
Suppose a utility company has $N$ data centers, each responsible for managing smart meter data from a set of prosumers, amounted $M_i$. At the $i$-th data center, a local model $f(\theta_i)$ is deployed and trained on its corresponding private dataset $\mathcal{D}_i$, where each sample pair $(X_i^d, \mathbf{y}^d_i)$ is drawn from $\mathcal{D}_i$. The local model $f(\theta_i)$ generates a prediction $\hat{\mathbf{y}}^d_i=f(\theta_i; X^d_i)$, which approximates the true label $\mathbf{y}^d_i$.
All data centers have the same objective to improve the performance by minimizing the empirical risk on their respective local datasets. For the $i$-th data center, the empirical risk can be formulated below:
\begin{align}
    \mathcal{F}_i:=\mathbb{E}_{(X_i^d,\mathbf{y}_i^d) \sim \mathcal{D}_i} \quad\mathcal{L}\big[f(\theta_i;X_i^d),\mathbf{y}_i^d\big],
\end{align}
where $\mathcal{L}$ is the loss function of PV disaggregation task to qualify the gap between model predictions $\hat{\mathbf{y}}$ and ground truth $\mathbf{y}$. The primary objective of distributed PV disaggregation is to personalize the local model parameters for each data center to minimize the empirical risk $\mathcal{F}_i$. 
The set of datasets for all data centers is denoted as $\mathcal{D} = \{\mathcal{D}_i\}_{i=1}^{N}$. Therefore, the training process aims to find a set of optimal local model parameters $\Theta^* = \{\theta^*_i\}^{N}_{i=1}$ as defined below:
\begin{align}
    \Theta^* = \mathop{\arg \min}\limits_{\theta_1,\dots,\theta_N} \sum_{i=1}^{N} \frac{|\mathcal{D}_i|}{|\mathcal{D}|} \mathcal{F}_i.
\end{align}

\section{proposed Privacy-Preserving Distributed PV Disaggregation Framework}\label{Sec:M}

In this section, the proposed privacy-preserving distributed PV disaggregation framework is explained, for addressing statistical heterogeneity using a PFL paradigm.
First, the feature engineering across multiple variables including the accessible net load readings and solar irradiance indicators, is introduced to provide richer information for the PV disaggregation model to learn more comprehensive PV generation patterns. 
Following this, the architecture of the PV disaggregation local model based on DL is explained, consisting of three key components: variate-centric embedding, transformer blocks, and an output layer. This design aims to improve PV disaggregation accuracy by generating representational vectors that capture temporal features and cross-variate dependencies across net load and solar irradiance indicators.
Finally, the adaptive PFL framework is outlined, which balances generalization and personalization for each data center by leveraging selective local model aggregation based on local PV conditions.

\subsection{Multivariate Feature Engineering on PV Disaggregation Factors}
In the local end, each data center is responsible for performing PV disaggregation for PV prosumers by processing recent time-series data. Due to privacy constraints, each data center only has access to net load readings from each prosumer, while PV system details, such as panel size and model specifications, are unavailable. To account for this limitation, external weather data is incorporated, specifically solar irradiance indicators, including Direct Horizontal Irradiance (DHI), Global Horizontal Irradiance (GHI), and Direct Normal Irradiance (DNI), which are highly related to PV generation and promisingly beneficial to improve PV disaggregation accuracy.

For each prosumer managed by the $i$-th data center, a sliding window contains the recent $L^{\text{Window}}$ days of data, sampled every half-hour, yielding a total of 48 time steps per day, and $48\times L^{\text{Window}}$ time steps per window.
For the $d$-th day, the input data of the $j$-th prosumer includes net load readings and the three irradiance metrics, i.e., DHI, GHI, and DNI, denoted as:
\begin{align}
X_{i,j}^d=\begin{bmatrix}(\mathbf{x}_{i,j}^{\text{Net},d-L^{\text{Window}}+1};\dots;\mathbf{x}_{i,j}^{\text{Net},d})\\(\mathbf{x}_{i,j}^{\text{DHI},d-L^{\text{Window}}+1};\dots;\mathbf{x}_{i,j}^{\text{DHI},d})\\(\mathbf{x}_{i,j}^{\text{DNI},d-L^{\text{Window}}+1};\dots;\mathbf{x}_{i,j}^{\text{DNI},d})\\(\mathbf{x}_{i,j}^{\text{GHI},d-L^{\text{Window}}+1};\dots;\mathbf{x}_{i,j}^{\text{GHI},d})
\end{bmatrix}
 \in \mathbb{R}^{4\times L^{\text{Window}}T}, \notag \\
i \in \{1,...,N\}, \quad d \in \{1,...,D\}, \quad j\in\{1,...,M_i\}.
\end{align}
The model's goal is to estimate the PV generation for the target day using the historical information. 

This sliding window approach enables the model to capture short-term dynamics in net load and irradiance data, allowing it to better understand the interactions between load and irradiance patterns, which is essential for accurate PV disaggregation.

\subsection{PV Disaggregation Model}
\begin{figure}[t]
\centering
\includegraphics[width=0.48\textwidth]{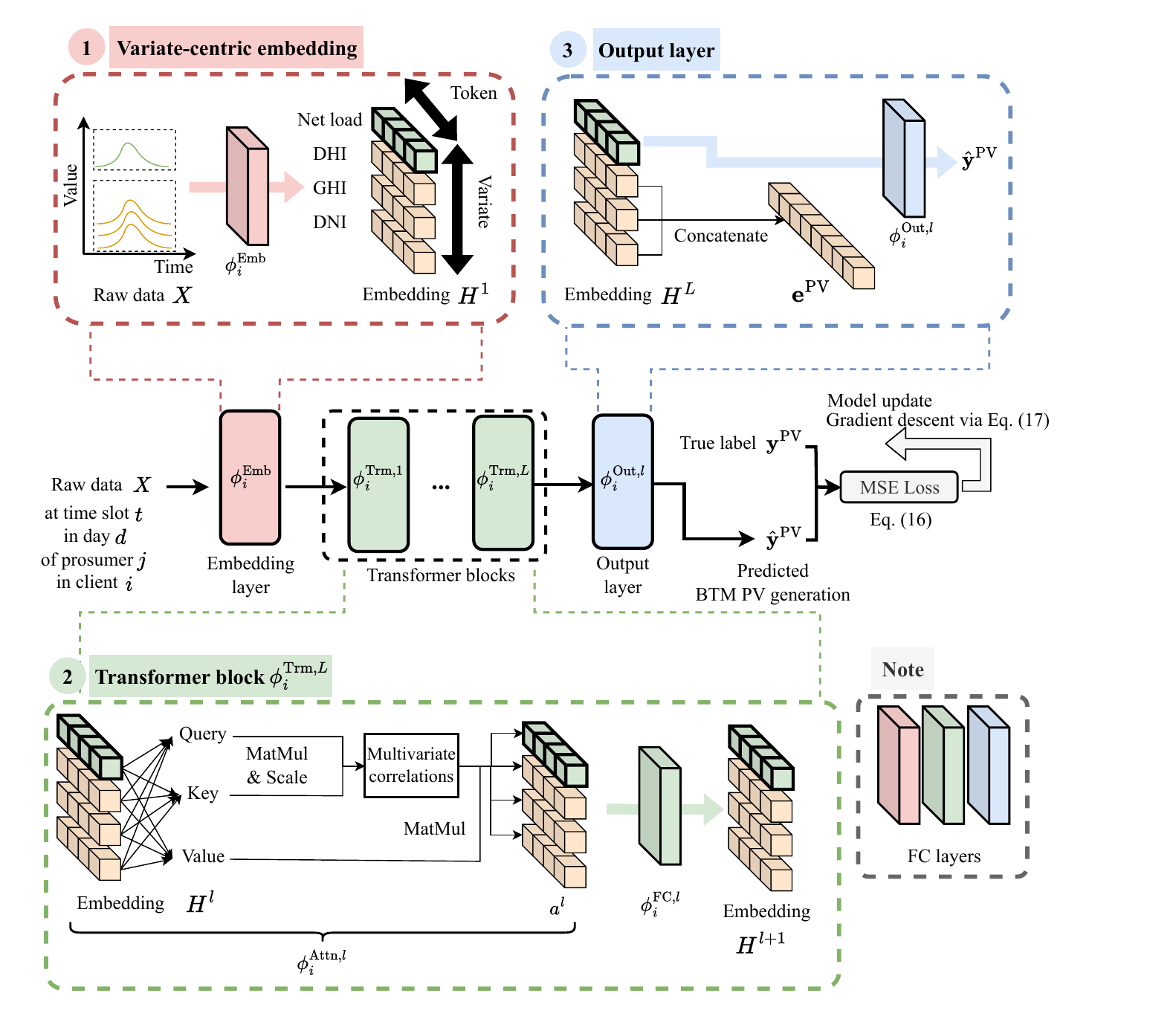}
\caption{The architecture of the proposed model for distributed PV disaggregation.}
\label{fig:model}
\end{figure}

The Transformer-based PV disaggregation model employs a variate-centric design with $L$ stacked Transformer blocks, capturing complex, long-range dependencies and interactions specifically across individual variates, e.g., net load, DHI, GHI, and DNI, rather than aggregating all variates per time step as in traditional Transformers. This approach enables the model to focus on cross-variate relationships and capture unique patterns for each variable over time, enhancing forecasting accuracy and robustness in PV disaggregation. Overall, the transformer-based PV disaggregation model shown in Fig.~\ref{fig:model} consists of three modules, including variate-centric embedding, transformer blocks, and an output layer. After feeding input variables forward through all these three modules, the model updates through gradient descent based on the designed loss function.

\subsubsection{Variate-Centric Embedding}

For each prosumer $j$ managed by data center $i$, the input time series data on day $d$ comprise net load and three irradiance metrics. Each variate is treated as a unique token to capture specific temporal patterns independently. The embedding layer before the first Transformer block, denoted as $\phi_i^{\text{Emb}}$, maps each variate's time series into a lower-dimensional representation, forming variate tokens that are concatenated into a matrix:
\begin{align}\label{Eq: emb} H^{1,d}_{i,j} = \begin{bmatrix} \phi_i^{\text{Emb}}(\mathbf{x}_{i,j}^{\text{Net},d}) \\ \phi_i^{\text{Emb}}(\mathbf{x}_{i,j}^{\text{DHI},d}) \\ \phi_i^{\text{Emb}}(\mathbf{x}_{i,j}^{\text{DNI},d}) \\ \phi_i^{\text{Emb}}(\mathbf{x}_{i,j}^{\text{GHI},d}) \end{bmatrix}, \end{align}
where $1$ indicates this matrix is the initial input matrix of the following transformer blocks.

\subsubsection{Transformer Blocks}

After processing the input variables by $\phi^{\text{Emb}}_i$, $L$ stacked Transformer blocks are utilized to process the embedded input data, capturing complex dependencies among variables and across time steps. Each Transformer block consists of a self-attention layer and a fully-connected (FC) layer. 
Briefly, the architecture of the $l$-th block is denoted as:
\begin{align}\label{Eq: trm}
    \phi^{\text{Trm},l}_{i} = \big[\phi^{\text{Attn},l}_{i};\phi^{\text{FC},l}_{i}\big].
\end{align}

The self-attention mechanism is adopted to capture dependencies between different variates and across time steps by computing attention scores among all input embeddings.
In block $l$, the input embeddings are denoted as $H^{l}_{i,j} \in \mathbb{R}^{4 \times d^{\text{Emb}}}$, and $d^{\text{Emb}}$ is the embedding dimension.
The queries $Q^{l}_{i,j}$, keys $K^{l}_{i,j}$, and values $V^{l}_{i,j}$ are computed by applying learned linear transformations to the input embeddings:
\begin{align} Q^{l}_{i,j} &= \phi^{\text{Attn},l}_{i}(H^{l,d}_{i,j}), \\ K^{l}_{i,j} &= \phi^{\text{Attn},l}_{i}(H^{l,d}_{i,j}), \\ V^{l}_{i,j} &= \phi^{\text{Attn},l}_{i}(H^{l,d}_{i,j}). \end{align}
The scaled dot-product attention is computed as:
\begin{align}\label{Eq: att}
\mathbf{a}^{l,d}_{i,j} = \text{Softmax}\left[ \frac{Q^{l}_{i,j} (K^l_{i,j})^{\top}}{\sqrt{d_k}} \right] V^{l}_{i,j}, \end{align}
where $d_k$ is the dimension of $K$.

Next, the attention is passed through a position-wise FC layer, which consists of two FC layers with a ReLU activation in between to get the embedding vector for next transformer block $\phi^{\text{Trm},l+1}_i$:
\begin{align} H^{l+1,d}_{i,j} = \phi^{\text{FC},l}_i(\mathbf{a}^{l,d}_{i,j}). \end{align}
The two-layer design introduces non-linearity to the model, enabling it to capture complex relationships among the variables.

\subsubsection{Output Layer and Update}
After processing through all $L$ Transformer blocks, the final representations $H^L_{i,j}$ of the four variables are obtained:
\begin{align}
    H^{L,d}_{i,j} = [\mathbf{h}^{\text{Net},L,d}_{i,j};\mathbf{h}^{\text{DHI},L,d}_{i,j};\mathbf{h}^{\text{DNI},L,d}_{i,j};\mathbf{h}^{\text{GHI},L,d}_{i,j}].
\end{align}
As for the model output, the net load embedding vector in $H^L_{i,j}$ is projected by a FC layer $\phi^{\text{Out}}_i$ to predict the PV generation for the target day:
\begin{align}\label{Eq: out}
\hat{\mathbf{y}}^{\text{PV},d}_{i,j} = \phi^{\text{Out}}_i(\mathbf{h}^{\text{Net},L,d}_{i,j}) \in \mathbb{R}^{T}, \end{align}
where $T = 48$ corresponds to the half-hour intervals of the target day.
The model of data center $i$ is trained by minimizing the Mean Squared Error (MSE) loss between the predicted PV generation and the true PV generation values for all samples in $\mathcal{D}_i$. The loss function is defined as:
\begin{align}\label{Eq: loss compute}
\mathcal{L}(\theta_i) = \frac{1}{M_i}\frac{1}{D}\frac{1}{T} \sum_{j=1}^{M_i}\sum_{d=1}^D\sum_{t=1}^T \left( \hat{y}^{\text{PV},d}_{i,j,t} - y^{\text{PV},d}_{i,j,t} \right)^2. \end{align}
The model parameters $\theta_i$ of data center $i$ are updated using gradient descent, which minimizes the loss function by adjusting the parameters in the direction of the negative gradient. The update rule is given by:
\begin{align}\label{Eq: loss back} \theta_i \leftarrow \theta_i - \eta_i \nabla_{\theta_i} \mathcal{L}(\theta_i), \end{align}
where $\eta_i$ is the learning rate of data center $i$. This iterative optimization enables the model to learn effective PV generation disaggregation patterns from historical data without requiring explicit PV system specifications.

On the other hand, the embedding vectors of the three PV indicators on the day $d$ are concatenated as one vector, and will be averaged among all prosumers in the most recent days to get a PV condition embedding vector which represents the most recent PV condition of the region monitored by data center $i$. The calculation will be discussed in detail in Section~\ref{Adaptive PFL Framework}.

\subsection{Adaptive PFL Framework}\label{Adaptive PFL Framework}
The PV disaggregation model in Fig.~\ref{fig:model} serves as the local model within the PFL framework. Multiple local models perform local training while leveraging the PFL framework's communication mechanism for model updates.
To balance generalization and personalization in PFL for PV disaggregation, a model-splitting mechanism is adopted, which divides each data center's local model $\theta_i$ into two distinct parts: the \textbf{base} consisting of lower layers and the \textbf{head} consisting of higher layers. Since the lower layers of DL models capture more generalized information compared to higher layers~\cite{NIPS2014_375c7134}, this design allows each data center to leverage shared global knowledge while personalizing its model based on local conditions.
Besides, solar irradiance embeddings derived from local irradiance data are utilized to adjust the influence of the global base model based on the similarity of local weather patterns to the global context, thus refining the balance between shared knowledge and local specificity.

Assume there are $R$ iterations for the PFL communication. At iteration $r$, according to Eq. (\!\!~\ref{Eq: emb}), Eq. (\!\!~\ref{Eq: trm}), and Eq. (\!\!~\ref{Eq: out}), the local model of data center $i$ is denoted as:
\begin{align}
    \theta_{i,r} = [\phi^{\text{Emb}}_{i,r}; \phi^{\text{Trm},1}_{i,r}; ...;\phi^{\text{Trm},L}_{i,r}; \phi^{\text{Out}}_{i,r}].
\end{align}

Then, for model splitting, a base model and a head model are defined as follows:
\begin{enumerate}
    \item \textbf{Base Model} $\theta^{\text{Base}}_{i,r} = [\phi^{\text{Emb}}_{i,r}; \phi^{\text{Trm},1}_{i,r}; \dots; \phi^{\text{Trm},L}_{i,r}]$: This part of the model captures generalized features by processing data through the embedding and Transformer layers. These layers learn broad patterns that are likely shared across regions, such as general relationships between energy load and weather conditions.
    \item \textbf{Head Model} $\theta^{\text{Head}}_{i,r} = [\phi^{\text{Out}}_{i,r}]$: The head model contains only the projection layer, which learns fine-grained, region-specific information necessary for accurate PV generation estimation in each unique environment.
\end{enumerate}
The model splitting can be denoted as:
\begin{align}
    \theta_{i,r} = [\theta^{\text{Base}}_{i,r}; \theta^{\text{Head}}_{i,r}].
\end{align}

Besides, a solar irradiance embedding vector $\mathbf{e}^{\text{PV}}_{i,r}$ is calculated to represent the PV condition of the region monitored by data center $i$. By averaging the DHI, DNI, and GHI embeddings in $\{H^{L,d}_{i,j}\}_{d=D^{\text{Rec}}_r}^D$ among each prosumer $j$ in the most recent days $D^{\text{Rec}}_r$ of iteration $r$, the embedding vector is obtained:
\begin{align}
    \mathbf{h}^{\text{PV},d}_{i,j} &= [\mathbf{h}^{\text{DHI},L,d}_{i,j},\mathbf{h}^{\text{DNI},L,d}_{i,j},\mathbf{h}^{\text{GHI},L,d}_{i,j}],\\
    \mathbf{e}^{\text{PV}}_{i,r} &= \frac{1}{M_i}\frac{1}{D-D^{\text{Rec}}_r} \sum_{j=1}^{M_i}\sum_{d=D^{\text{Rec}}_r}^D \mathbf{h}^{\text{PV},d}_{i,j}.
\end{align}

Next, the entire process of the proposed PFL framework is summarized.
At the beginning, each data center $i$ trains its local model on its local data $\mathcal{D}_i$ and updates all the parameters of $\theta_{i,r}$, as well as calculating the solar irradiance embedding vector $\mathbf{e}^{\text{PV}}_{i,r}$.

After local training, each data center $i$ shares its base model $\theta^{\text{Base}}_{i,r}$ and solar irradiance embedding vector $\mathbf{e}^{\text{PV}}_{i,r}$ with the server. The head model $\theta^{\text{Head}}_{i,r}$, which contains region-specific information, remains local to each data center. 

The server aggregates the base models and solar irradiance embedding vectors separately across data centers using a weighted averaging approach based on data volumes owned by data centers:
\begin{align}
    \theta^{\text{Base},\text{G}}_r = \sum_{i=1}^{N} \frac{|\mathcal{D}_i|}{|\mathcal{D}|} \theta^{\text{Base}}_{i,r}, \\
    \mathbf{e}^{\text{PV},\text{G}}_{r} = \sum_{i=1}^{N} \frac{|\mathcal{D}_i|}{|\mathcal{D}|} \mathbf{e}^{\text{PV}}_{i,r},
\end{align}
where $|\mathcal{D}|$ is the total data volume of all data centers. 
This aggregation produces a global base model $\theta^{\text{Base}, \text{G}}_r$ and a global solar irradiance embedding vector $\mathbf{e}^{\text{PV},\text{G}}_{r}$, which integrates generalized patterns across data centers without compromising individual data privacy.

Once the global base model $\theta^{\text{Base},\text{G}}_r$ and global solar irradiance embedding vector $\mathbf{e}^{\text{PV},\text{G}}_{r}$ are obtained, they are sent back to each data center. 

Locally, each data center $i$ first calculates the Cosine similarity between $\mathbf{e}^{\text{PV}}_{i,r}$ and $\mathbf{e}^{\text{PV},\text{G}}_{r}$ and maps it into $[0,1]$ to get $\lambda_i$:
\begin{align}
    S_{i,r} &= \frac{\mathbf{e}^{\text{PV}}_{i,r} \cdot \mathbf{e}^{\text{PV},\text{G}}_{r}}{||\mathbf{e}^{\text{PV}}_{i,r}||  ||\mathbf{e}^{\text{PV},\text{G}}_{r}||} \in [-1,1], \label{Eq: similarity}\\
    \lambda_{i,r} &= \frac{S_{i,r}+1}{2} \in [0,1],\label{Eq: lambda}
\end{align}
where ``$\cdot$'' is dot product operation.
When the similarity is higher, indicating the PV condition of the data center $i$ is similar to the major PV condition among all data centers, $\lambda_i$ is larger to allow the global model participate more into the local model.
Then, data center $i$ combines the global base model with its local base model using $\lambda_i$, and concatenates its local head model to create an aggregated local model:
\begin{align}
    \hat{\theta}^{\text{Base}}_{i,r} &= \lambda_{i,r} \theta^{\text{Base},\text{G}}_r + (1-\lambda_{i,r}) \theta^{\text{Base}}_{i,r},\label{Eq: aggregate}\\
    \hat{\theta}_{i,r+1} &= [\hat{\theta}^{\text{Base}}_{i,r}, \theta^{\text{Head}}_{i,r}].\label{Eq: aggregate 2}
\end{align}
This personalized model allows each data center to apply the generalized knowledge from the global base while preserving local insights captured by its specific head model.

To further adapt the global base model to each region's specific conditions, each data center $i$ conducts local training on $\hat{\theta}_{i,r+1}$ using its local dataset according to Eq. (\!\!~\ref{Eq: loss back}), and obtains $\theta_{i,r+1}$ for next-round communication.

\begin{figure}[t]
\centering
\includegraphics[width=0.48\textwidth]{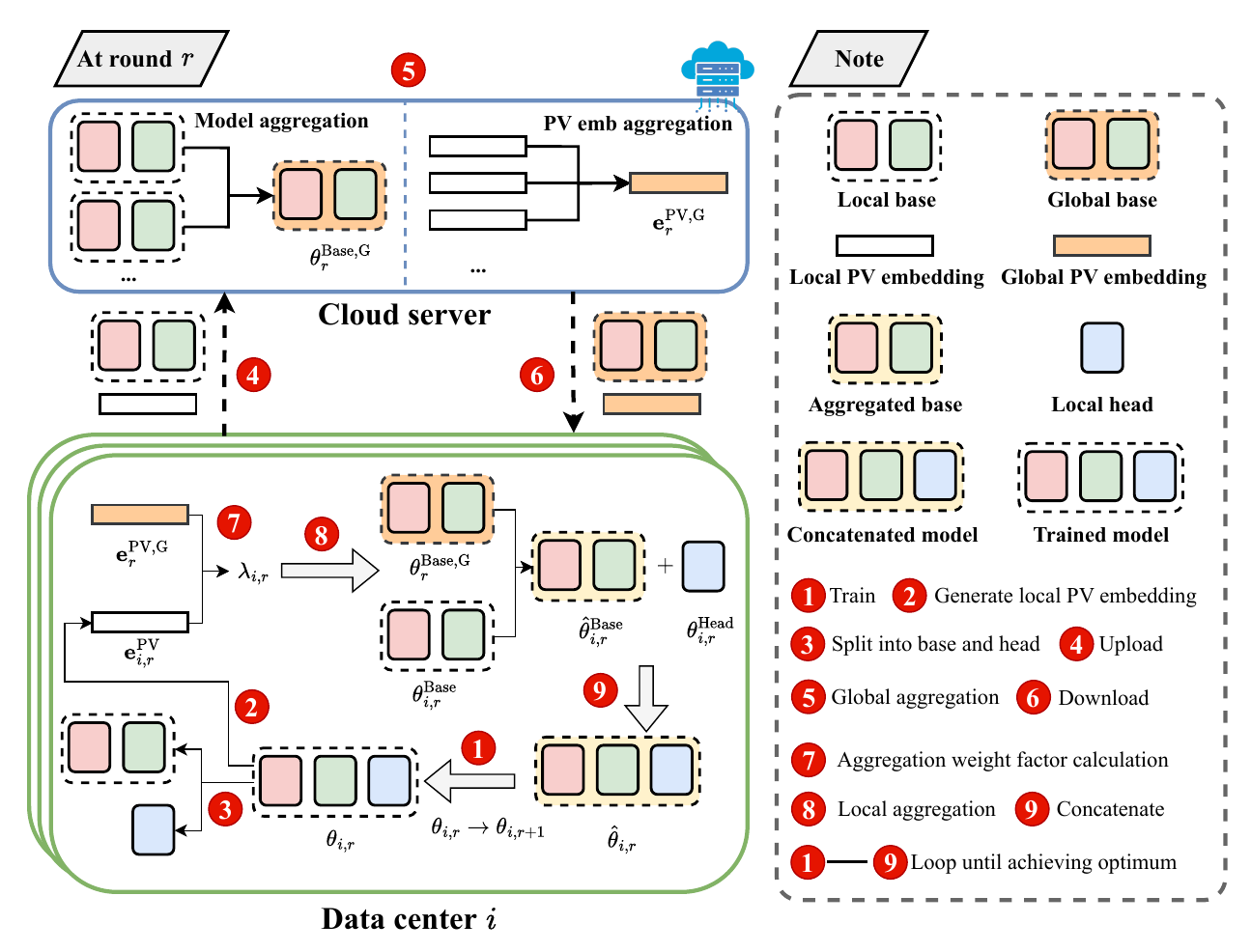}
\caption{Illustration of the proposed PFL framework for distributed PV disaggregation.} 
\label{fig:FL}
\end{figure}

The proposed architecture of PFL framework for distributed PV disaggregation is depicted in Fig.~\ref{fig:FL}. The PV disaggregation model in Fig.~\ref{fig:model} serves as the local model of the PFL framework. Multiple local models perform local training while leveraging the PFL framework's communication mechanism for model updates. At the $r$-th round of PFL communication, each data center $i$ trains its local PV disaggregation model $\theta_{i,r}$, using private data. The local model learns to disaggregate PV generation based on the net load and weather data. A local solar irradiance embedding, $e^{\text{PV}}_{i,r}$, is generated from the trained model, representing key regional PV characteristics. The local model $\theta_{i,r}$ is split into a local base model $\theta^{\text{Base}}_{i,r}$ and a local head model $\theta^{\text{Head}}_{i,r}$ to enable personalization. Then, each data center uploads its local base model $\theta^{\text{Base}}_{i,r}$ and local solar irradiance embedding $e^{\text{PV}}_{i,r}$ to the cloud server for global aggregation. The cloud server performs model aggregation on the received base models from all data centers, forming the global base model $\theta^{\text{Base,G}}_{i,r}$. Similarly, local solar irradiance embeddings are aggregated to form the global solar irradiance embedding $e^{\text{PV,G}}_{i,r}$. The cloud server sends the global base model $\theta^{\text{Base,G}}_{i,r}$ and global solar irradiance embedding $e^{\text{PV,G}}_{i,r}$ back to each data center. Each data center computes the local aggregation weighting factor $\lambda_{i,r}$, which determines how much influence the global base model should have on the local model. This factor is computed based on both the global and local solar irradiance embeddings. The global base model $\theta^{\text{Base,G}}_{i,r}$ is locally aggregated with the local base model $\theta^{\text{Base}}_{i,r}$ using $\lambda_{i,r}$, generating an updated base model $\hat{\theta}^{\text{Base}}_{i,r}$. The updated base model $\hat{\theta}^{\text{Base}}_{i,r}$ is concatenated with the unchanged local head model $\theta^{\text{Head}}_{i,r}$, forming the final local model $\hat{\theta}_{i,r}$. The iterative process continues until local models converge.

\renewcommand{\baselinestretch}{1.1} 
\begin{algorithm}[t] \footnotesize
\caption{Privacy-Preserving Distributed PV Disaggregation PFL Framework}\label{alg1}
\begin{algorithmic}[1]
\STATE {\bfseries Input:} Local data $\{\mathcal{D}_i\}_{i=1}^N$; Local models $\{\theta_i\}_{i=1}^N$; Learning rates $\{\eta_i\}_{i=1}^N$; Number of communication iterations $R$; Number of Transformer layers $L$. 
\STATE {\bfseries Output:} Optimized personalized PV disaggregation model $\{\theta^*_i\}_{i=1}^N$. 
\STATE {\bfseries Initialization:}
\STATE Each data center $i$ initializes its model $\theta_{i,0}$. 
\STATE Server initializes the global model $\theta^{\text{Base}, \text{G}}_{0}$. 
\STATE Initialize weighting factor of each data center $i$ $\lambda_{i,r}$ as 0.5.
\STATE {\bfseries FL communication:}
\STATE \textbf{for} communication iteration $r = 1$ to $R$ \textbf{do}
\STATE \hspace{0.3cm} {\bfseries Clients:}
\STATE \hspace{0.3cm} \textbf{for} each data center $i$ in parallel \textbf{do} \hfill $\blacktriangleright$ \textbf{Local Model Update}
\STATE \hspace{0.6cm} \textbf{if} $r>1$, calculate weighting factor $\lambda_{i,r}$. \hfill $\blacktriangleright$ \circled{7}
\STATE \hspace{0.6cm} \textbf{else} continue.
\STATE \hspace{0.6cm} Obtain aggregated local model $\hat{\theta}_{i,r}$ following $\lambda_i$. \hfill $\blacktriangleright$ \circled{8}, \circled{9}
\STATE \hspace{0.6cm} \textbf{for} each prosumer $j$ \textbf{do} \hfill $\blacktriangleright$ \textbf{Local Training}
\STATE \hspace{0.9cm} \textbf{for} each day $d$ \textbf{do}
\STATE \hspace{1.2cm} Obtain input $X_{i,j}^d$.
\STATE \hspace{1.2cm} Compute embedding $H^1_{i,j}$ according to Eq. (\!\!~\ref{Eq: emb}).
\STATE \hspace{1.2cm} Obtain $H^L_{i,j}$ according to Eq. (\!\!~\ref{Eq: att}).
\STATE \hspace{1.2cm} Predict PV generation $\hat{\mathbf{y}}^{\text{PV},d}_{i,j}$ according to Eq. (\!\!~\ref{Eq: out}).
\STATE \hspace{0.9cm} \textbf{end for}
\STATE \hspace{0.6cm} \textbf{end for}
\STATE \hspace{0.6cm} Compute loss $\mathcal{L}(\hat{\theta}_{i,r})$ using Eq. (\!\!~\ref{Eq: loss compute}). 
\STATE \hspace{0.6cm} Update model parameters $\theta_{i,r} \leftarrow \hat{\theta}_{i,r} - \eta_i \nabla{\hat{\theta}_{i,r}} \mathcal{L}(\hat{\theta}_{i,r})$. \hfill $\blacktriangleright$ \circled{1}
\STATE \hspace{0.6cm} Calculate local solar irradiance embedding vector $\mathbf{e}^{\text{PV}}_{i,r}$. \hfill $\blacktriangleright$ \circled{2}
\STATE \hspace{0.6cm} Split model into base $\theta^{\text{Base}}_{i,r}$ and head $\theta^{\text{Head}}_{i,r}$. \hfill $\blacktriangleright$ \circled{3}
\STATE \hspace{0.6cm} Send $\theta^{\text{Base}}_{i,r}$ and $\mathbf{e}^{\text{PV}}_{i,r}$ to the global server. \hfill $\blacktriangleright$ \circled{4}
\STATE \hspace{0.3cm} \textbf{end for}
\STATE \hspace{0.3cm} {\bfseries Server:} \hfill $\blacktriangleright$ \circled{5}
\STATE \hspace{0.3cm} Aggregate base models to obtain global base model $\theta^{\text{Base},\text{G}}_r$.
\STATE \hspace{0.3cm} Aggregate local solar irradiance embeddings to obtain global embedding $\mathbf{e}^{\text{PV},G}_r$.
\STATE \hspace{0.3cm} Send $\theta^{\text{Base},\text{G}}_r$ and $\mathbf{e}^{\text{PV},G}_r$ back to each data center. \hfill $\blacktriangleright$ \circled{6}
\STATE \hspace{0.3cm} \textbf{end for}
\STATE \textbf{end for}
\STATE \textbf{return} $\theta^*_1, \theta^*_2, \dots, \theta_N^*$. 
\end{algorithmic}
\end{algorithm} 
\renewcommand{\baselinestretch}{1.0}

Additionally, Algorithm~\ref{alg1} presents the training process of the improved privacy-preserving distributed PV disaggregation framework. The process includes: \textbf{Initialization} of local models at each data center and the global base model at the server (Lines 3-6); \textbf{Federated Learning Communication Iterations} (Lines 7-33); and \textbf{Finalization}. After completing all communication iterations, the optimized personalized PV disaggregation models $\{\theta^*_i\}_{i=1}^N$ are obtained for all data centers (Line 34).
Next, an analysis of the computational and communication efficiency is provided. The local PV embedding is computed by first projecting the input data onto a $d^{\text{Emb}}$-dimensional space, then processing these embeddings through multiple Transformer blocks with a total of $d_k$ neurons, and finally aggregating daily embeddings across prosumers. 
While each step has its own computational cost, when considering dataset- and model-specific constants, the overall complexity simplifies to $\mathcal{O}(d^{\text{Emb}} \cdot d_k)$.
In terms of communication overhead, each client transmits only the base model and a compact PV embedding to the server, keeping the additional communication cost minimal. Consequently, the overall communication overhead is lower than that of traditional federated learning methods, such as FedAvg, while still effectively enhancing local adaptation under statistical heterogeneity.

\section{Experimental Study and Results}\label{Sec:E}
In this section, an experimental study is conducted to evaluate the effectiveness of the proposed method. First, the dataset used in the experiments is analyzed. Next, an outline of the performance metrics and baseline methods for comparison is introduced. Finally, the results are analyzed, focusing on the performance of the local model, the effectiveness of the PFL framework, and the impact of a new data center participation.

\subsection{Dataset Description}
\begin{figure}[t]
    \centering   
    \includegraphics[width=.45\textwidth]{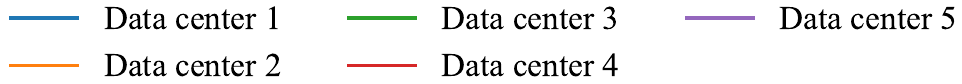}
    \subfigure[DHI]{
        \includegraphics[width=.22\textwidth]{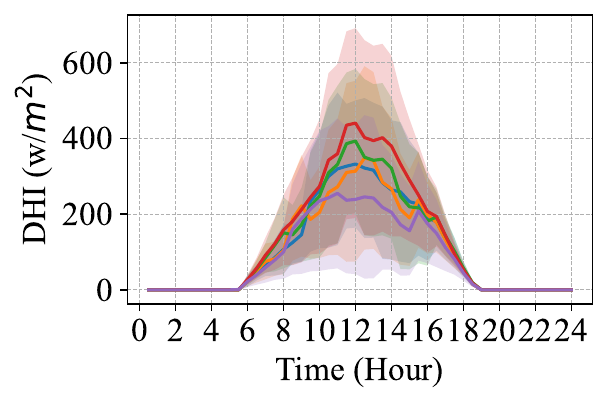}  
        \label{Fig:nrel_ana_dhi}
        }
    \subfigure[DNI]{
        \includegraphics[width=.22\textwidth]{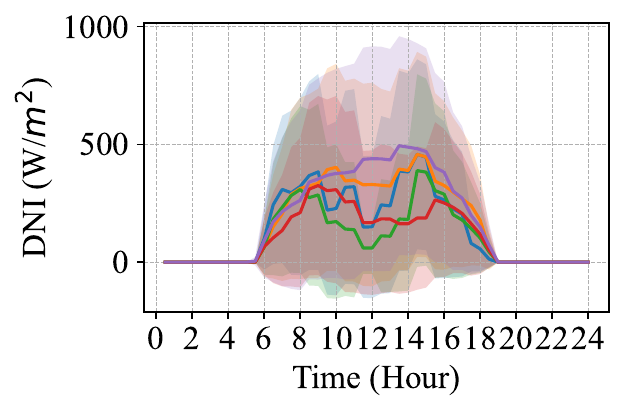} 
        \label{Fig:nrel_ana_dni}
        }
    \subfigure[GHI]{
        \includegraphics[width=.22\textwidth]{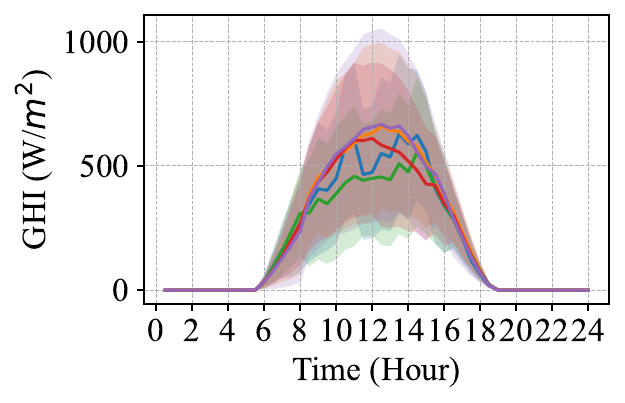}  
        \label{Fig:nrel_ana_ghi}
        }
    \subfigure[Net Load]{
        \includegraphics[width=.21\textwidth]{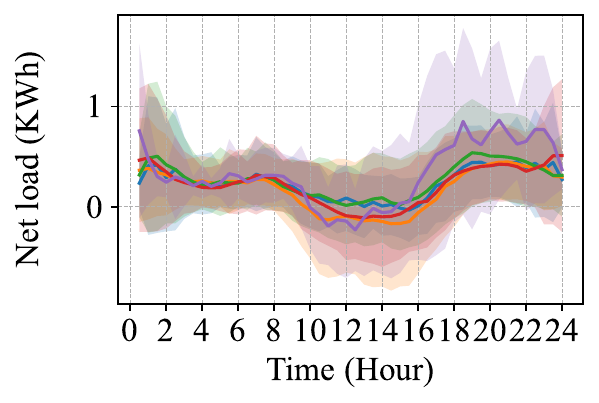}  
        \label{Fig:aus_ana}
        }
    \caption{Mean-variance visualization of raw solar irradiance and raw net load data for five data centers in one week.}
    \label{Fig:all_ana}
\end{figure}

The dataset employed in these experiments is the Solar Home Electricity Data~\cite{ratnam2017residential} provided by Ausgrid's electricity network, comprising three years of half-hourly electricity data for 300 randomly selected solar homes from July 1, 2010, to June 30, 2013. This dataset encompasses two main categories of consumption: 1) general and controlled load consumption, representing total household electricity usage excluding PV generation; and 2) gross generation, recording the total electricity produced by the solar PV systems independently of household consumption. Additionally, weather data for these 300 solar homes were sourced from the National Solar Radiation Database~\cite{NREL}, which offers comprehensive meteorological data. For this study, three principal solar radiation metrics are utilized: global horizontal irradiance (GHI), direct normal irradiance (DNI), and diffuse horizontal irradiance (DHI).

To optimize data transmission costs, electricity consumption and generation data from neighboring solar homes are typically stored in the same data center. 300 solar homes are categorized into five data centers based on the geographical distribution.
Fig.~\ref{Fig:all_ana} provides a mean-variance visualization of the daily solar irradiance and net load data for each data center over a randomly selected week. Each solid line in the figure represents the mean of daily irradiance or net load samples for a data center. The shaded regions around each line indicate the variance in irradiance and net load samples in each data center.
In Fig.~\ref{Fig:nrel_ana_dhi}, Fig.~\ref{Fig:nrel_ana_dni}, and Fig.~\ref{Fig:nrel_ana_ghi}, while the overall trends are similar, differences in the midday peak amplitudes are apparent among the data centers. Certain data centers exhibit higher variance, particularly at specific times of day, reflecting increased or decreased variability in solar irradiance due to factors such as cloud cover and atmospheric conditions. These variances reveal \textit{geographical heterogeneity}, where each data center's unique location and climate result in distinct irradiance patterns that influence PV generation.
Additionally, the shading in Fig.~\ref{Fig:aus_ana} varies throughout the day, indicating significant differences in variance across data centers, which further reflect the \textit{heterogeneity of prosumer behavior}, as group differences in electricity usage and device operation lead to temporal fluctuations in net load patterns.

\subsection{Performance Metrics and Benchmark Methods}

To assess the accuracy of the proposed framework, three evaluation metrics are employed: Mean Absolute Error (MAE), Root Mean Square Error (RMSE), and the coefficient of determination (R$^2$). These metrics provide a comprehensive analysis of the model's performance in terms of both absolute error and variability explanation.

The MAE is formulated as below:
\begin{align}
    \text{MAE} = \frac{1}{T}\sum_{t=1}^T |\hat{ \mathbf{y}}_t-\mathbf{y}_t|,
\end{align}
where $T$ represents the total number of time points for a day, $\mathbf{y}_t$ is the true value at time $t$, and $\hat{\mathbf{y}}_t$ is the predicted value at time $t$.
The RMSE is formulated as:
\begin{align}
    \text{RMSE} =\sqrt{ \frac{1}{T}\sum_{t=1}^T (\hat{\mathbf{y}}_t-\mathbf{y}_t)^2},
\end{align}
and the R$^2$ is formulated as:
\begin{align}
    \text{R}^2 =\frac{\sum_{i=1}^{T}(\hat{\mathbf{y}}_t-\bar{\mathbf{y}}_t)^2}{\sum_{i=1}^{T}(\mathbf{y}_t-\bar{\mathbf{y}}_t)^2}=1-\frac{\sum_{i=1}^{T}(\mathbf{y}_t-\hat{\mathbf{y}}_t)^2}{\sum_{i=1}^{T}(\mathbf{y}_t-\bar{\mathbf{y}}_t)^2},
\end{align}
where $\bar{\mathbf{y}}_t$ is the mean of the true value at time $t$.

In the proposed framework, both a novel local model and a PFL framework are integrated. To comprehensively evaluate the performance of each component, separate comparisons between the local model and popular deep learning models are conducted, as well as between this PFL approach and established FL methods.
For the local model, centralized training evaluations against several baselines are conducted, including MLP, LSTM~\cite{hochreiter1997long}, Transformer~\cite{10.5555/3295222.3295349}, Reformer~\cite{Kitaev2020Reformer}, Informer~\cite{Zhou_Zhang_Peng_Zhang_Li_Xiong_Zhang_2021}, Autoformer~\cite{wu2021autoformer}, and DLinear~\cite{Zeng_Chen_Zhang_Xu_2023}.
For the PFL framework, three baselines are compared: Local-only, FedAvg~\cite{pmlr-v54-mcmahan17a} as a traditional FL framework, and Ditto~\cite{pmlr-v139-li21h} as a PFL framework. In the Local-only approach, each data center independently trains a model using only its local data, without inter-center communications, as a baseline for fully decentralized learning. All distributed training methods are implemented using the proposed local model to ensure a consistent comparison.

\begin{table}[t]
\renewcommand{\arraystretch}{1.1}
\centering
\caption{Performance Results of Centralized Training Evaluations}
\label{tab:centralized}
\begin{tabular}{cccc}
\hline\hline
Method   & MAE (KWh)    & RMSE (KWh)   & R$^2$     \\ \hline
MLP               & 0.0773          & 0.1384          & 0.4830          \\
LSTM              & 0.0706          & 0.1169          & 0.6852          \\
Transformer       & 0.0643          & 0.1076          & 0.7433          \\
Reformer          & 0.0630          & 0.1060          & 0.7537          \\
Informer          & 0.0636          & 0.1075          & 0.7444          \\
Autoformer        & 0.0653          & 0.1087          & 0.7379          \\
DLinear           & 0.0675          & 0.1141          & 0.7071          \\
\textbf{Proposed} & \textbf{0.0621} & \textbf{0.1038} & \textbf{0.7757} \\ \hline\hline
\end{tabular}
\end{table}

\subsection{Result Analysis}

\subsubsection{Comparison of Local Model Performance}
As shown in Table~\ref{tab:centralized}, the proposed model outperforms all baselines across all three metrics, demonstrating superior accuracy, lower error deviations, and greater explanatory ability. 
The results demonstrate that the proposed local model effectively distinguishes between electricity consumption and generation data by leveraging the relationship with solar irradiance. Furthermore, it demonstrates the ability to capture temporal patterns and internal relationships between net load and irradiance data, generating informative embeddings.
Among the baseline methods, Reformer, Informer, and Transformer exhibit competitive performance, with Reformer performing consistently close to the proposed model in three metrics. In contrast, MLP and LSTM exhibit weaker performance, particularly in R$^2$, revealing their limitations in capturing complex patterns for PV disaggregation.

\begin{figure}[t]
\centering
\includegraphics[width=0.48\textwidth]{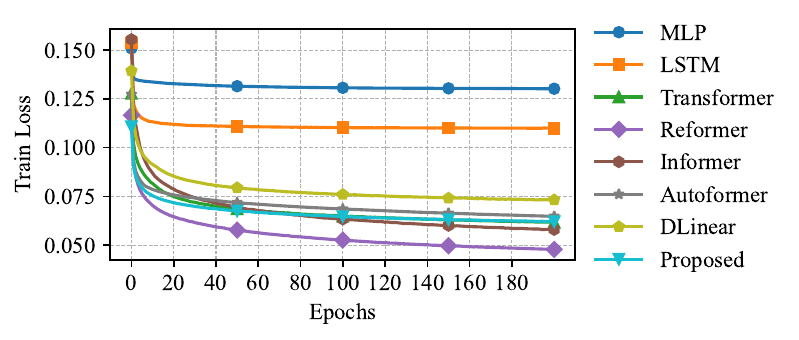}
\caption{The learning curves of centralized training models.} 
\label{fig:cen_loss}
\end{figure}

\begin{figure}[t]
\centering
\includegraphics[width=0.468\textwidth]{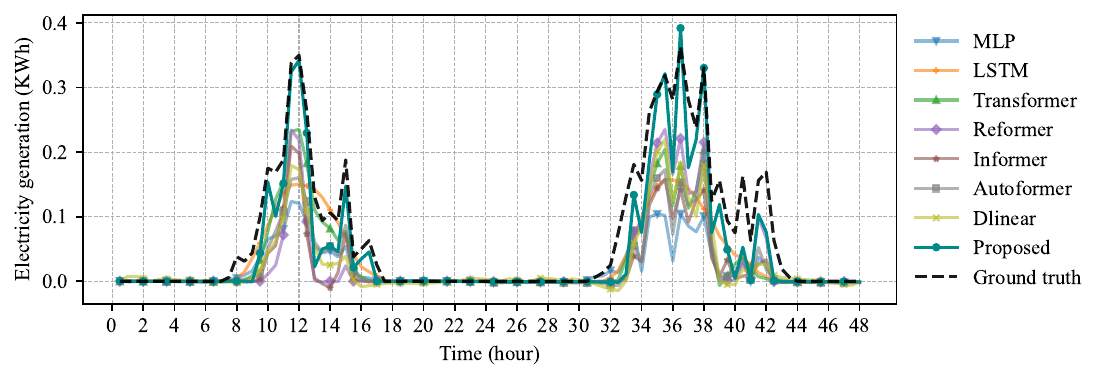}
\caption{PV disaggregation estimation results of centralized training models for solar home \#74 from July 23, 2012 to July 24, 2012.}
\label{fig:testdata_trend_cen_2}
\end{figure}

Furthermore, the learning curves in Fig.~\ref{fig:cen_loss} illustrate the training losses based on MSE of various models. Compared to the baselines, the proposed model demonstrates a faster convergence rate. It is also notable that while Reformer and Informer achieve relatively low training losses, their performance on the test set is not optimal, indicating potential overfitting or weaker generalization capability.

Moreover, the PV generation disaggregation results for a randomly selected solar home over two consecutive days as shown in Fig.~\ref{fig:testdata_trend_cen_2}. The proposed model demonstrates a strong alignment with the ground truth, capturing both the peak generation around midday and the fluctuations during the afternoon more accurately than the baseline models. Other models, such as Reformer and Informer, tend to underestimate peak values, leading to less precise profiles. Interestingly, models like MLP and LSTM appear to capture the general trend but lack precision during peak hours, suggesting limitations in capturing complex temporal dependencies.

Overall, the proposed model's close alignment with the actual generation curve reveals its ability to capture complex temporal dependencies and nonlinear relationships between net load, irradiance, and PV generation. This strong performance indicates its ability to extract irradiance features related to PV generation, which accurately represent local PV conditions for selecting beneficial global knowledge.

\begin{table*}[t]
\renewcommand{\arraystretch}{1.2}
\scriptsize
\centering
\caption{Performance Results of Distributed Training Evaluations}
\label{tab:pfl}
\begin{tabular}{ccccccccccccc}
\hline\hline
              & \multicolumn{3}{c|}{\textbf{Proposed}}                                  & \multicolumn{3}{c|}{Local-only}                                & \multicolumn{3}{c|}{FedAvg}                                    & \multicolumn{3}{c}{Ditto}                \\ \cline{2-13} 
              & MAE & RMSE & \multicolumn{1}{c|}{R$^2$} & MAE & RMSE & \multicolumn{1}{c|}{R$^2$} & MAE & RMSE & \multicolumn{1}{c|}{R$^2$} & MAE & RMSE & R$^2$ \\ \hline
Data center 1 & \textbf{0.0592} & \textbf{0.0945} & \multicolumn{1}{c|}{\textbf{0.6877}} & 0.0606    & 0.0962   & \multicolumn{1}{c|}{0.6661}   & 0.0613  & 0.0971  & \multicolumn{1}{c|}{0.6609}  & 0.0598  & 0.0952 & 0.6746 \\
Data center 2 & \textbf{0.0760} & \textbf{0.1350} & \multicolumn{1}{c|}{\textbf{0.8077}} & 0.0792    & 0.1389   & \multicolumn{1}{c|}{0.7785}   & 0.0799  & 0.1391  & \multicolumn{1}{c|}{0.7800}  & 0.0782  & 0.1376 & 0.7903 \\
Data center 3 & \textbf{0.0612} & \textbf{0.1007} & \multicolumn{1}{c|}{\textbf{0.7257}} & 0.0631    & 0.1033   & \multicolumn{1}{c|}{0.6888}   & 0.0639  & 0.1049  & \multicolumn{1}{c|}{0.6782}  & 0.0625  & 0.1026 & 0.6918 \\
Data center 4 & \textbf{0.0633} & \textbf{0.1095} & \multicolumn{1}{c|}{\textbf{0.7778}} & 0.0650    & 0.1120   & \multicolumn{1}{c|}{0.7508}   & 0.0656  & 0.1125  & \multicolumn{1}{c|}{0.7586}  & 0.0640  & 0.1105 & 0.7678 \\ \hline
Data center 5 & \textbf{0.0871} & \textbf{0.1292} & \multicolumn{1}{c|}{\textbf{0.7483}} & 0.1104    & 0.1514   & \multicolumn{1}{c|}{0.5569}   & 0.0954  & 0.1395  & \multicolumn{1}{c|}{0.6264}  & 0.0913  & 0.1355 & 0.6569 \\ \hline
\end{tabular}
\end{table*}

\subsubsection{Evaluation of PFL Framework Effectiveness}
In this subsection, experiments involve the first four data centers, while Data center 5 remains to be examined in the following subsection. It is worth noting that the data quantity across the first four data centers is relatively uniform, with no apparent skew or unbalancedness. As shown in the Table~\ref{tab:pfl}, the proposed PFL framework consistently achieves the lowest MAE and RMSE values across four data centers, along with higher R$^2$ values compared to other methods. This shows that the proposed method performs most closely approximates the performance of a centralized training approach. The Local-only method underperforms relative to PFL frameworks, including the proposed method and Ditto, suggesting that integrating knowledge from multiple centers enhances local model performance. In contrast, FedAvg shows the weakest performance, implying difficulty in handling heterogeneous data distributions across data centers. This limitation may arise because FedAvg averages local model updates from all centers, potentially neglecting the distribution of data features specific to each center. While Ditto performs relatively closer to the proposed method, it still falls short, demonstrating the superiority of the proposed knowledge-sharing and personalization strategies in addressing this regression task.

\begin{figure}[t]
\centering
\includegraphics[width=0.48\textwidth]{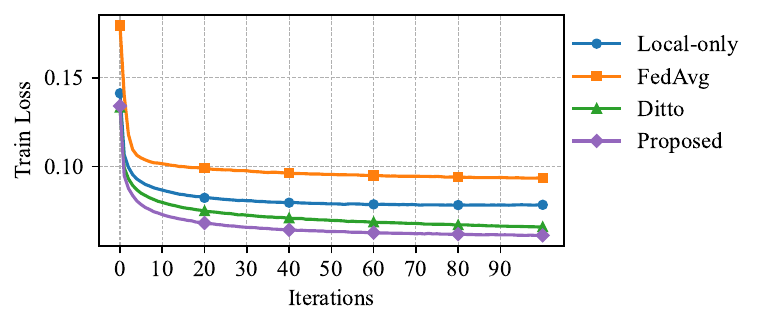}
\caption{The learning curves of distributed training frameworks.} 
\label{fig:pfl_loss}
\end{figure}

\begin{figure}[t]
\centering
\includegraphics[width=0.468\textwidth]{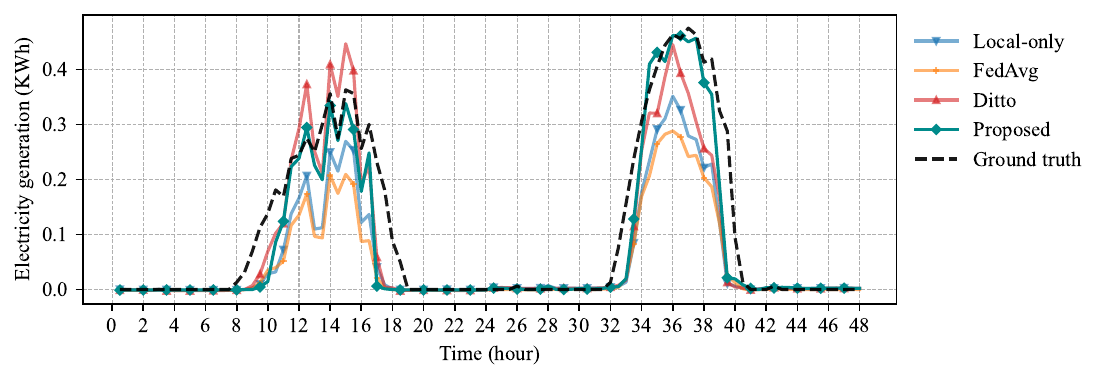}
\caption{PV disaggregation estimation results of distributed training frameworks for solar home \#231 from June 15, 2012 to June 16, 2012.}
\label{fig:testdata_trend_fl}
\end{figure}

Besides, the training loss of the four distributed frameworks of 100 iterations, measured by MSE, is illustrated in the Fig.~\ref{fig:pfl_loss}. The proposed PFL framework exhibits a relatively fast convergence rate, achieving the lowest training loss throughout the iterations. Ditto also demonstrates competitive convergence behavior, while Local-only reaches a higher final loss than Ditto, indicating that solely training models without joint knowledge sharing yields less effective results. FedAvg starts with a relatively high initial loss and converges slowly, likely due to the weight divergence mentioned in \cite{9743558}, which hinders its globally shared model from reaching a true global optimum.

In addition, Fig.~\ref{fig:testdata_trend_fl} illustrates the PV generation disaggregation performance, comparing four distributed learning frameworks. The proposed PFL framework demonstrates the closest fit to the ground truth, capturing both the magnitude and timing of the peaks more accurately than the other approaches. Ditto also performs relatively well, with a closer fit to the peaks than Local-only, though it still exhibits some deviations. In contrast, FedAvg shows a less accurate fit, particularly around the peak generation hours.

\subsubsection{New Data Center Participation}

In addition to the notable differences in data distribution across existing centers, utility companies may also establish new data centers over time as part of their ongoing operations. 
New data centers often face the challenge of limited historical data, a form of quantity skew. This subsection examines the scenario where, following the completion of the initial training process, a new data center-Data center 5, is introduced. Data center 5 has significantly fewer data samples, possessing only 8\% of the data volume of other centers, providing an opportunity to assess the framework's robustness and adaptability when handling centers with new scarce data resources.

Compared to previous training, the introduction of Data center 5 requires only a few training iterations for the distributed framework to exhibit a clear convergence trend.
The experimental results, as shown in the last row of Table~\ref{tab:pfl}, demonstrate that the proposed method achieves the best performance across all metrics. This suggests that the proposed approach can effectively generalize and adapt to new data centers by sharing generalized knowledge and leveraging similar PV conditions, even in scenarios of data scarcity.
The Local-only approach, which lacks cross-center knowledge sharing, performs the worst across all metrics, with a notably low R$^2$ value of 0.5569. This outcome shows the limitations of training solely on restricted local data without incorporating external knowledge, resulting in underfitting. Similarly, while FedAvg improves upon the Local-only method, its performance remains suboptimal, as its global model-averaging approach struggles to capture the unique data features of the new data center. 
Ditto's performance, while better than Local-only and FedAvg, still falls short of the proposed framework, suggesting that Ditto's local model personalization is less effective than the proposed framework's approach to knowledge sharing and local adaptation.

\section{Conclusion}\label{Sec:C}
In this paper, a novel privacy-preserving distributed PV disaggregation framework is proposed for prosumers with PV systems under statistical heterogeneity. Based on the PFL paradigm, the proposed method balances the need for generalization and personalization by employing a two-level framework with a transformer-based local PV disaggregation model and a novel local aggregation mechanism. Extensive experiments on real-world datasets demonstrate the effectiveness of the method. The results show that the tailored design of the local model using the Transformer-based architecture, along with the training process in the proposed PFL framework, contributes to high-accuracy PV disaggregation in such distributed learning scenarios. It provides a scalable solution to addressing data privacy, statistical heterogeneity, and personalized adaptation through hierarchical model splitting and local-global aggregation. 

While the proposed framework is designed for PV disaggregation, such a paradigm holds potential to be extended to other energy disaggregation tasks with appropriate modifications. The authors plan to explore more energy disaggregation tasks in future work.
Another research direction is to explore semi-supervised and unsupervised learning approaches to reduce the reliance on labeled data, improving model adaptability in data-limited or privacy-sensitive contexts. 
Real-world smart meter data are often noisy, incomplete, or faulty, and addressing these challenges by developing more robust methods is critical for future research.

\bibliography{reference}
\bibliographystyle{IEEEtran}

\begin{IEEEbiography}[{\includegraphics[width=1in,height=1.25in,clip,keepaspectratio]{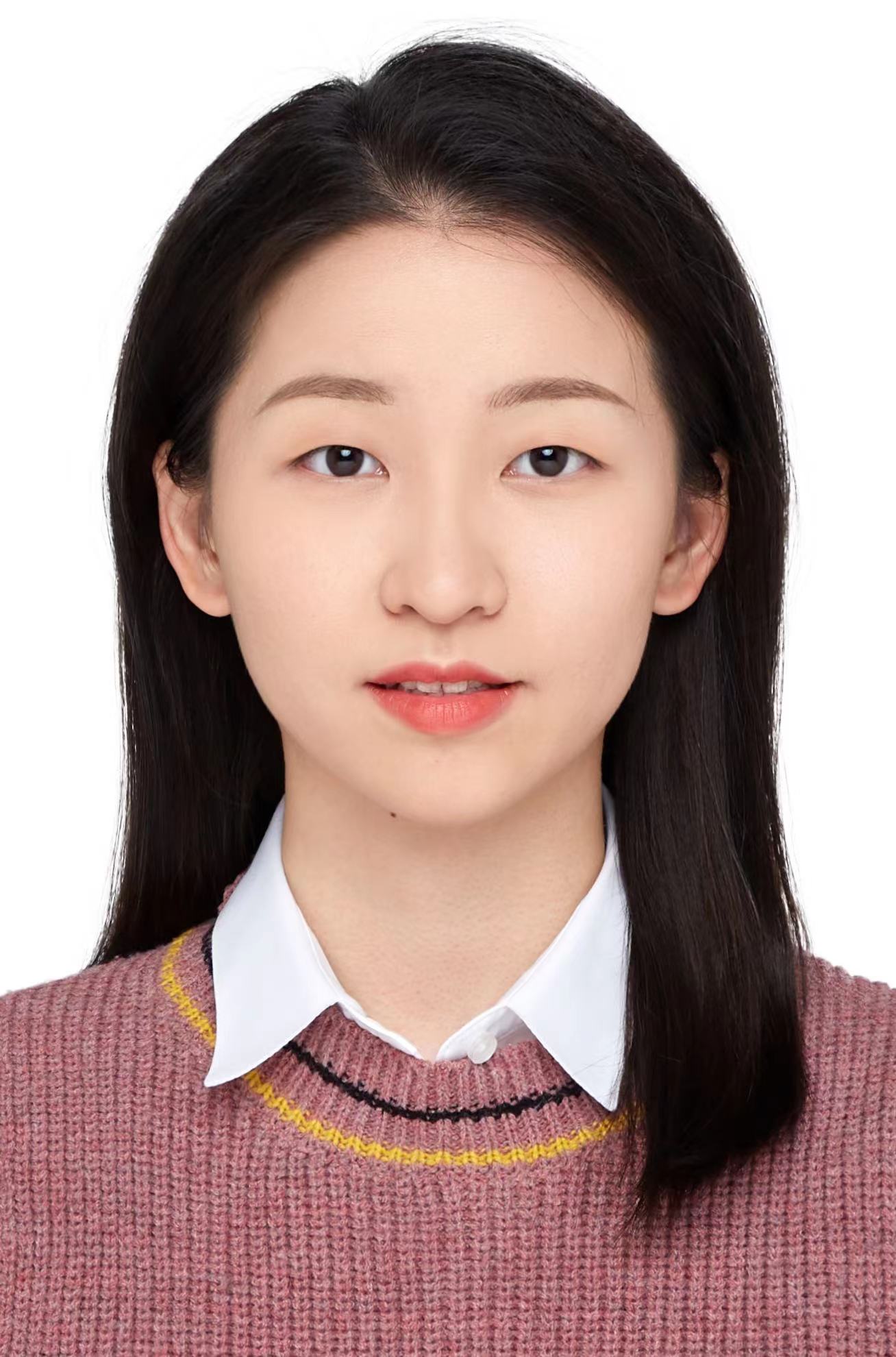}}]{Xiaolu Chen}
received a B.E. degree from the Department of Computer Science and Technology, at the University of Electronic Science and Technology of China, in 2022. She is currently pursuing M.S. degree with the Department of Computer Science and Technology, at the University of Electronic Science and Technology of China. Her main research interests include deep learning, federated learning, and smart grid.
\end{IEEEbiography}

\begin{IEEEbiography}
[{\includegraphics[width=1in,height=1.25in,clip,keepaspectratio]{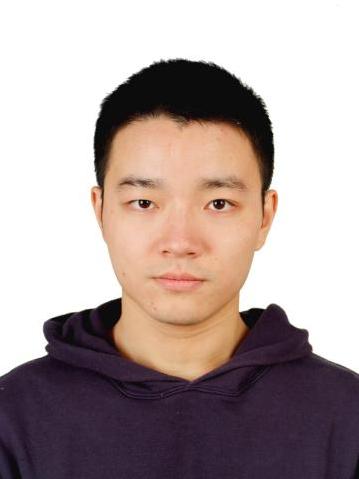}}] 
{Chenghao Huang}
received the B.E. degree in software engineering and M.S. degree in computer science and engineering from University of Electronic Science and Technology of China (UESTC). He is currently pursuing the Ph.D. degree with the Faculty of Information Technology, Monash University. His research interests include deep learning, federated learning, reinforcement learning and smart grid.
\end{IEEEbiography}

\begin{IEEEbiography}[{\includegraphics[width=1in,height=1.25in,clip,keepaspectratio]{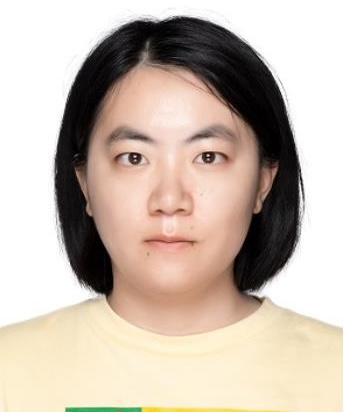}}]{Yanru Zhang}
(S’13-M’16) received the B.S. degree in electronic engineering from the University of Electronic Science and Technology of China (UESTC) in 2012, and the Ph.D. degree from the Department of Electrical and Computer Engineering, University of Houston (UH) in 2016. She worked as a Postdoctoral Fellow at UH and the Chinese University of Hong Kong successively. She is currently a Professor with the Shenzhen Institute for Advanced Study and School of Computer Science, UESTC. Her current research involves game theory, machine learning, and deep learning in network economics, Internet and applications, wireless communications, and networking. She received the Best Paper Award at IEEE HPCC 2022, DependSys 2022, ICCC 2017, and ICCS 2016. 
\end{IEEEbiography}

\begin{IEEEbiography}
[{\includegraphics[width=1in,height=1.25in,clip,keepaspectratio]{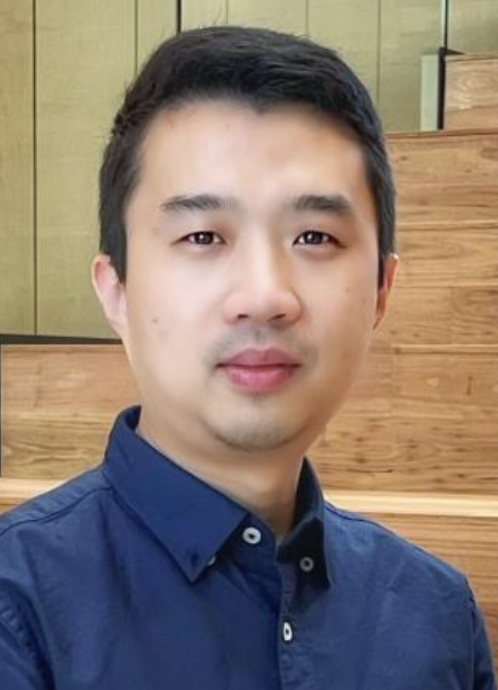}}]{Hao Wang} (M'16) received his Ph.D. in Information Engineering from The Chinese University of Hong Kong, Hong Kong, in 2016. He was a Postdoctoral Research Fellow at Stanford University, Stanford, CA, USA, and a Washington Research Foundation Innovation Fellow at the University of Washington, Seattle, WA, USA. He is currently a Senior Lecturer and ARC DECRA Fellow in the Department of Data Science and AI, Faculty of IT, Monash University, Melbourne, VIC, Australia. His research interests include optimization, machine learning, and data analytics for power and energy systems.
\end{IEEEbiography}

\vfill

\end{document}